\documentclass[final]{elsarticle}
\usepackage{times,amsmath,amssymb,subfigure,url,multirow,color}
\usepackage[flushleft]{threeparttable}
\usepackage[lined,ruled,commentsnumbered]{algorithm2e}
\allowdisplaybreaks

\begin{document}

\begin{frontmatter}
\title{Active Stacking for Heart Rate Estimation}

\author[HUST1,HUST2]{Dongrui Wu\corref{CA}}\ead{drwu@hust.edu.cn}
\author[SEU]{Feifei Liu}\ead{feifeiliu1987@gmail.com}
\author[SEU]{Chengyu Liu\corref{CA}}\cortext[CA]{Corresponding authors}\ead{chengyu@seu.edu.cn}

\address[HUST1]{Key Laboratory of Image Processing and Intelligent Control\\ (Huazhong University of Science and Technology), Ministry of Education, China}
\address[HUST2]{School of Artificial Intelligence and Automation,\\ Huazhong University of Science and Technology, Wuhan, China}

\address[SEU]{State Key Laboratory of Bioelectronics, School of Instrument Science and Engineering,\\ Southeast University, Nanjing, China}

\begin{abstract}
Heart rate estimation from electrocardiogram signals is very important for the early detection of cardiovascular diseases. However, due to large individual differences and varying electrocardiogram signal quality, there does not exist a single reliable estimation algorithm that works well on all subjects. Every algorithm may break down on certain subjects, resulting in a significant estimation error. Ensemble regression, which aggregates the outputs of multiple base estimators for more reliable and stable estimates, can be used to remedy this problem. Moreover, active learning can be used to optimally select a few trials from a new subject to label, based on which a stacking ensemble regression model can be trained to aggregate the base estimators. This paper proposes four active stacking approaches, and demonstrates that they all significantly outperform three common unsupervised ensemble regression approaches, and a supervised stacking approach which randomly selects some trials to label. Remarkably, our active stacking approaches only need three or four labeled trials from each subject to achieve an average root mean squared estimation error below three beats per minute, making them very convenient for real-world applications. To our knowledge, this is the first research on active stacking, and its application to heart rate estimation.
\end{abstract}

\end{frontmatter}

\section{Introduction}

Cardiovascular diseases are the leading cause of human death. According to the World Health Organization \cite{WHOCD}, cardiovascular diseases take 17.9 million lives every year, accounting for 31\% of all global deaths. Electrocardiogram (ECG) is very useful in early detection of cardiovascular diseases. Recent years have witnessed rapid developments of wearable ECG systems for continuous ECG monitoring \cite{Liu2019}. In such systems, real-time accurate heart rate estimation is critical to cardiovascular disease detection and treatment \cite{Khamis2016}. Unfortunately, ECG from these wearable systems generally has poor quality due to bad electrode contact, body movements, and various noise \cite{Li2008c}. As a result, traditional heart rate estimation algorithms, which mainly considered clinic quality ECG signals, may have difficulty on the wearable ECG systems \cite{Liu2018}. Additionally, even when the ECG signal quality is satisfactory, due to large individual differences, there may not exist a single heart rate estimation algorithm that works well on all subjects. This paper considers how to use advanced machine learning approaches to cope with these problems.

Ensemble regression \cite{Zhou2012} has been frequently used to improve the estimation performance, by integrating multiple base estimators. More specifically, we assume $M$ base estimators are used to estimate the heart rates of $N$ ECG trials from a particular subject. According to whether labeled training data are available or not, there are two types of ensemble regression approaches:
\begin{enumerate}
\item \emph{Unsupervised ensemble regression}, where no labeled ECG trials are available. The simplest, maybe also the most frequently used, unsupervised ensemble regression approach is to take the average of the $M$ base estimators. However, as it will be shown later in this paper, because of individual differences, this approach does not work well on heart rate estimation.
\item \emph{Supervised ensemble regression}, where some labeled ECG trials are available. Some sophisticated supervised ensemble regression approaches \cite{Buhlmann2010}, e.g., bagging \cite{Breiman1996}, boosting \cite{Freund1997a,Friedman2001}, random forests \cite{Breiman2001}, etc, require a relatively large number of labeled data. The simplest supervised ensemble regression approach, which also does not require too many labeled data, may be stacking \cite{Breiman1996b}, i.e., the final estimator is a weighted average of the base estimators, where the weights are computed from the labeled training data. Again, as it will be shown later in this paper, because of individual differences, it is very challenging, if not impossible, to find a set of weights that fits all subjects. Usually some subject-specific labeled ECG trials must be obtained, based on which a subject-specific ensemble regression approach can then be designed to achieve a high estimation accuracy.
\end{enumerate}

Intuitively, supervised ensemble regression would outperform unsupervised ensemble regression, if high-quality subject-specific labeled ECG trials can be acquired. Generally, the more such trials there are, the higher the estimation accuracy will be. However, for practical considerations, we'd like to minimize the number of subject-specific labeled ECG trials, as labeling each such trial requires an expert to visually examine and count the number of QRS waves in the ECG trial, which is both tedious and labor-intensive. So, it is desirable to be able to reduce the number of subject-specific labeled ECG trials.

Active learning \cite{Settles2009} is a popular and effective approach for this purpose. It deliberately selects a small number of most beneficial trials from the $N$ unlabeled trials to label, so that a model trained from these labeled trials can achieve the best possible performance. Our previous research has demonstrated the outstanding performance of active learning in both classification \cite{drwuRSVP2016,drwuTNSRE2016} and regression \cite{drwuSAL2019,drwuMTALR2019,drwuiGS2019} tasks, in a variety of application domains. However, to our knowledge, no one has integrated stacking and active learning for heart rate estimation.

This paper proposes four novel active stacking approaches for estimator aggregation, which integrate active learning for regression (ALR) \cite{drwuSAL2019,drwuMTALR2019,drwuiGS2019} and stacking. The idea is to use ALR to select a small number of most beneficial unlabeled trials, query an expert for their outputs, and then train a stacking model on them. We demonstrate their outstanding performances on heart rate estimation from ECG signals on 95 subjects: our active stacking approaches only need three or four labeled ECG trials from each subject to achieve an average root mean squared estimation error below three beats per minute, making them very practical for real-world applications.

The remainder of this paper is organized as follows: Sections~\ref{sect:ER} and \ref{sect:ALR} introduce three ensemble regression approaches and four ALR approaches, respectively, which will be used in our study. Section~\ref{sect:AS} proposes four active stacking approaches. Section~\ref{sect:experiments} compares their performances on heart rate estimation from ECG signals. Finally, Section~\ref{sect:conclusion} draws conclusion.

\section{Ensemble Regression} \label{sect:ER}

Three simple yet popular ensemble regression approaches are introduced in this section. We assume there are $N$ ECG trials from a particular subject, and $M$ base estimators have been applied to each trial to estimate the heart rates $\mathbf{x}_n=[x_{1,n},...,x_{M,n}]^T$ beats per minute (bpm), $n=1,...N$. We would like to use ensemble regression to aggregate each $\mathbf{x}_n$ for a more accurate estimate, $\hat{y}_n$.

\subsection{Average} \label{sect:average}

The simplest ensemble regression approach is to take the average of the $M$ base estimators, i.e.,
\begin{align}
\hat{y}_n=\frac{1}{M}\sum_{m=1}^M x_{m,n},\quad n=1,...,N \label{eq:average}
\end{align}
Note that (\ref{eq:average}) does not need any labeled training trials, i.e., it's a completely unsupervised ensemble regression approach.

\subsection{Median} \label{sect:median}

Another simple ensemble regression approach is to take the median of the $M$ base estimators, i.e.,
\begin{align}
\hat{y}_n=\underset{m}{\mathrm{median}} (x_{m,n}),\quad n=1,...,N \label{eq:median}
\end{align}
Note that (\ref{eq:median}) does not need any labeled training trials, either, i.e., it's also a completely unsupervised ensemble regression approach.

\subsection{Stacking}

Stacking is a supervised ensemble regression approach \cite{Breiman1996b}. Assume among the $N$ ECG trials, $K$ have been labeled, i.e., their reference heart rates $\{y_k\}_{k=1}^K$ are known. Then, stacking trains a regression model $\hat{y}_n=f(\mathbf{x}_n)$ from these $K$ trials. Ridge regression and linear support vector regression (SVR) were used in this paper.

A ridge regression model is:
\begin{align}
\hat{y}_n=\mathbf{w}^T\mathbf{x}_n+b,\quad n=1,...,N \label{eq:rr}
\end{align}
where $b$ and $\mathbf{w}=[w_1,...,w_M]^T$ are obtained from minimizing the following objective function:
\begin{align}
g(b,\mathbf{w})=\sum_{k=1}^K\left(y_k-\hat{y}_k\right)^2+\lambda \mathbf{w}^T\mathbf{w}
\end{align}
in which $\lambda=0.01$.

A linear SVR model \cite{Vapnik1995} can also be represented by (\ref{eq:rr}), but now $b$ and $\mathbf{w}$ minimize the following objective function:
\begin{align}
g(b,\mathbf{w})&=\frac{1}{2}\mathbf{w}^T\mathbf{w}+C\sum_{k=1}^K \epsilon_k \\
s.t.\quad &|y_k-\hat{y}_k|\le \epsilon_k,\quad \epsilon_k\ge 0
\end{align}
in which $C=1$.

\section{Active Learning for Regression (ALR)} \label{sect:ALR}

This section introduces four ALR approaches. The first two are unsupervised, whereas the last two are supervised.

Assume a subject has $N$ ECG trials, each with its heart rate estimates $\mathbf{x}_n$ from the $M$ base estimators, but initially none of these trials has a reference heart rate label. The goal of ALR is to optimally select $K$ trials to label, so that an accurate regression model can be constructed from them to estimate the heart rate for the remaining $N-K$ trials.

\subsection{GSx} \label{sect:GS}

Yu and Kim \cite{Yu2010} proposed a greedy sampling (GS) ALR approach, which selects the trials to label based entirely on their locations in the input space. Thus, it does not need any label information at all. However, the original GS approach did not explain how the first trial was selected. We \cite{drwuiGS2019} recently introduced GSx to accommodate this. GSx is essentially the same as GS, except that it also includes a strategy to select the first trial for labeling.

GSx selects the first trial as the one whose $\mathbf{x}_n$ is the closest to the centroid of all $N$ $\mathbf{x}_n$ (i.e., the one with the shortest mean distance to the remaining $N-1$ $\mathbf{x}_n$), and the remaining $K-1$ trials incrementally. In this way, the first selected trial is the most representative one in the $N$ trials.

Without loss of generality, assume the first $k$ trials $\{\mathbf{x}_l\}_{l=1}^k$ have already been selected. For each of the remaining $N-k$ unlabeled trials $\{\mathbf{x}_n\}_{n=k+1}^N$, GSx computes first its distance to each of the $k$ labeled trials:
\begin{align}
d_{nl}^{\mathbf{x}}=||\mathbf{x}_n-\mathbf{x}_l||,\quad l=1,...,k;\ n=k+1,...,N \label{eq:dnmx}
\end{align}
then $d_n^{\mathbf{x}}$, the shortest distance from $\mathbf{x}_n$ to all $k$ labeled trials:
\begin{align}
d_n^{\mathbf{x}}=\min_l d_{nl}^{\mathbf{x}},\quad n=k+1,...,N \label{eq:dnx}
\end{align}
and finally selects the trial with the maximum $d_n^{\mathbf{x}}$ to label.

In summary, GSx selects the first trial as the one closest to the centroid of the pool, and each subsequent trial located farthest away from all previously selected ones in the input space, to achieve the maximum diversity among the selected trials.

\subsection{RD}

We \cite{drwuSAL2019} recently proposed a representativeness-diversity (RD) approach for ALR. As its name suggests, it considers both representativeness and diversity in all trial selections. In contrast, GSx considers only representativeness in selecting the first trial, and only diversity in subsequent selections.

RD selects all $K$ trials simultaneously. It performs $k$-means ($k=K$) clustering on the $N$ unlabeled trials, and then selects from each cluster the trial closest to the cluster centroid for labeling. This selection strategy ensures representativeness, because each trial is a good representation of the cluster it belongs to. It also ensures diversity, because these $K$ clusters cover the full input space of $\mathbf{x}_n$, and they do not overlap.

As GSx, RD does not need any reference label information at all, so it is a completely unsupervised ALR approach.

\subsection{RD-EMCM}

RD only considers representativeness and diversity. However, as pointed out in \cite{drwuSAL2019}, informativeness is also an essential criterion in ALR. An RD-EMCM ALR approach was proposed in \cite{drwuSAL2019}, which considers also the informativeness through expected model change maximization (EMCM) \cite{Cai2013}.

RD-EMCM first uses RD to select $K_0=2$ trials, and queries for their reference labels. To select the next trial to label, it performs $k$-means ($k=K_0+1$) clustering on the $N$ trials, and identifies the largest cluster that does not already contain any labeled trial. It will then select the ($K_0+1$)th trial from this cluster. However, instead of selecting the one closest to its centroid, as in RD, now it uses EMCM to select the most informative trial to label. The details of EMCM are given next.

EMCM first uses all labeled trials to build a linear regression model (e.g., ridge regression, or linear SVR). Let its estimated heart rate for the $n$th trial be $\hat{y}_n$. EMCM then uses bootstrap to construct another $P$ linear regression models from the labeled trials. Let the $p$th model's estimated heart rate for the $n$th trial be $\hat{y}_n^p$. Then, for each unlabeled trials, EMCM computes \cite{Cai2013}
\begin{align}
g(\mathbf{x}_n)=\frac{1}{P}\sum_{p=1}^P\left\| (\hat{y}_n^p-\hat{y}_n)\mathbf{x}_n\right\|, \label{eq:EMCM}
\end{align}
and selects the trial with the maximum $g(\mathbf{x}_n)$ to label.

RD-EMCM is a supervised ALR approach, because it needs the reference labels to train the regression models in EMCM.

\subsection{iGS} \label{sect:iGS}

Improved greedy sampling (iGS) is an ALR approach proposed in \cite{drwuiGS2019}, which is supposed to improve GSx by considering also feature selection/weighting. It is a supervised ALR approach.

iGS first uses GSx to select the initial $K_0=2$ trials to label. Assume the first $k$ trials $\{\mathbf{x}_l\}_{l=1}^k$ have already been labeled with true heart rates $\{y_l\}_{l=1}^k$. For each of the remaining $N-k$ unlabeled trials $\{\mathbf{x}_n\}_{n=k+1}^N$, iGS computes:
\begin{align}
d_{nl}^{\mathbf{x}}&=||\mathbf{x}_n-\mathbf{x}_l|| \label{eq:dnmx}\\
d_{nl}^y&=|f(\mathbf{x}_n)-y_l| \label{eq:dnmy}\\
d_n^{\mathbf{x}y}&=\min_l d_{nl}^{\mathbf{x}}d_{nl}^y\label{eq:dnxy}
\end{align}
and then selects the trial with the maximum $d_n^{\mathbf{x}y}$, i.e., the trial located farthest away from all previously selected trials in both input and output spaces, to label.

\section{Active Stacking} \label{sect:AS}

Stacking requires some labeled trials, whereas ALR can optimally select a small number of trials to label. So, it's natural to integrate them for better performance. Four active stacking approaches are proposed next.

\subsection{AS-GSx}

AS-GSx integrates stacking and GSx. It uses GSx to select $K$ trials to query for their reference heart rates, and then checks if any base estimator has the same heart rate estimates as the reference for all $K$ selected trials. If yes, then for each of the remaining $N-K$ trials, the median of these base estimators is taken as its final estimate. Otherwise, it trains a linear SVR model from the $K$ labeled trials as the final stacking model.

The pseudo-code of AS-GSx is given in Algorithm~\ref{alg:GSx}.

\begin{algorithm}[htpb]
\KwIn{$N$ unlabeled trials, $\{\mathbf{x}_n\}_{n=1}^N$\;
\hspace*{10mm} $K$, the maximum number of labels to query.}
\KwOut{The stacking regression model $f(\mathbf{x})$.}
Set $Z=\{\mathbf{x}_n\}_{n=1}^N$, and $S=\emptyset$\;
Identify $\mathbf{x}'$, the trial closest to the centroid of $Z$\;
Move $\mathbf{x}'$ from $Z$ to $S$\;
Re-index the trial in $S$ as $\mathbf{x}_1$, and the trials in $Z$ as $\{\mathbf{x}_n\}_{n=2}^N$\;
\For{$k=1,...,K-1$}{
\For{$n=k+1,...,N$}{
Compute $d_n^{\mathbf{x}}$ in (\ref{eq:dnx})\;}
Identify the $\mathbf{x}'$ that has the largest $d_n^{\mathbf{x}}$\;
Move $\mathbf{x}'$ from $Z$ to $S$\;
Re-index the trials in $S$ as $\{\mathbf{x}_l\}_{l=1}^{k+1}$, and the trials in $Z$ as $\{\mathbf{x}_n\}_{n=k+2}^N$\;}
Query to label all $K$ trials in $S$\;
\uIf{There exist some base estimators which give identical estimates to the true labels in $S$}
{$f(\mathbf{x})$ is the median of these base estimator outputs\;}
\Else{Construct a linear SVR model $f(\mathbf{x})$ from $S$.}
\caption{The AS-GSx active stacking approach.} \label{alg:GSx}
\end{algorithm}

\subsection{AS-RD}

AS-RD integrates stacking and RD. It's almost identical to AS-GSx, except that GSx is replaced by RD as the ALR approach. Its pseudo-code is given in Algorithm~\ref{alg:RD}.

\begin{algorithm}[htpb] 
\KwIn{$N$ unlabeled trials, $\{\mathbf{x}_n\}_{n=1}^N$\;
\hspace*{10mm} $K$, the maximum number of labels to query.}
\KwOut{The stacking regression model $f(\mathbf{x})$.}
Perform $k$-means clustering on $\{\mathbf{x}_n\}_{n=1}^N$, where $k=K$\;
Select from each cluster the trial closest to its centroid, and query for its label\;
\uIf{There exist some base estimators which give identical estimates to the true labels for all $K$ trials}
{$f(\mathbf{x})$ is the median of these base estimator outputs\;}
\Else{Construct a linear SVR model $f(\mathbf{x})$ from the $K$ labeled trials.}
\caption{The AS-RD active stacking approach.} \label{alg:RD}
\end{algorithm}

\subsection{AS-RD-EMCM}

AS-RD-EMCM integrates stacking and RD-EMCM. It first uses RD-EMCM to select $K_0=2$ trials to query for their reference heart rates, and trains a linear SVR stacking model from them. This model can then be used in RD-EMCM to select the next trial to label, and the linear SVR stacking model is then updated. This process iterates until $K$ trials have been selected and labeled. Finally, we check if any base estimator has the same heart rate estimates as the reference for all $K$ selected trials. If yes, then for each of the remaining $N-K$ trials, the median of these base estimators is taken as the final estimate. Otherwise, we train a linear SVR model from the $K$ labeled trials as the final stacking model.

The pseudo-code of AS-RD-EMCM is given in Algorithm~\ref{alg:RD-EMCM}.

\begin{algorithm}[htpb] 
\KwIn{$N$ unlabeled trials, $\{\mathbf{x}_n\}_{n=1}^N$\;
\hspace*{10mm} $K$, the maximum number of labels to query.}
\KwOut{The stacking regression model $f(\mathbf{x})$.}
Perform $k$-means clustering on $\{\mathbf{x}_n\}_{n=1}^N$, where $k=2$\;
Select from each cluster the trial closest to its centroid, and query for its label\;
Construct a linear SVR model $f(\mathbf{x})$ from the two labeled trials\;
\For{$k=3,...,K$}{
{Perform $k$-means clustering on $\{\mathbf{x}_n\}_{n=1}^N$\;
Identify the largest cluster that does not already contain any labeled trial\;
Compute $g(\mathbf{x}_n)$ in (\ref{eq:EMCM}) for each trial in the above cluster\;
Select the trial with the maximum $g(\mathbf{x}_n)$ to label\;
Construct a linear SVR model $f(\mathbf{x})$ from the $k$ labeled trials\;}}
\uIf{There exist some base estimators which give identical estimates to the true labels for all $K$ trials}
{$f(\mathbf{x})$ is the median of these base estimator outputs\;}
\Else{Construct a linear SVR model $f(\mathbf{x})$ from the $K$ labeled trials.}
\caption{The AS-RD-EMCM active stacking approach.} \label{alg:RD-EMCM}
\end{algorithm}

\subsection{AS-iGS}

AS-iGS integrates stacking and iGS. It's almost identical to AS-RD-EMCM, except that RD-EMCM is replaced by iGS as the ALR approach. Its pseudo-code is given in Algorithm~\ref{alg:iGS}.

\begin{algorithm}[htpb] 
\KwIn{$N$ unlabeled trials, $\{\mathbf{x}_n\}_{n=1}^N$\;
\hspace*{10mm} $K$, the maximum number of labels to query.}
\KwOut{The stacking regression model $f(\mathbf{x})$.}
Set $Z=\{\mathbf{x}_n\}_{n=1}^N$, and $S=\emptyset$\;
Identify $\mathbf{x}'$, the trial closest to the centroid of $Z$\;
Move $\mathbf{x}'$ from $Z$ to $S$\;
Re-index the trial in $S$ as $\mathbf{x}_1$, and the trials in $Z$ as $\{\mathbf{x}_n\}_{n=2}^N$\;
\For{$n=2,...,N$}{
Compute $d_n^{\mathbf{x}}$ in (\ref{eq:dnx})\;}
Identify the $\mathbf{x}'$ that has the largest $d_n^{\mathbf{x}}$\;
Move $\mathbf{x}'$ from $Z$ to $S$\;
Re-index the trials in $S$ as $\{\mathbf{x}_l\}_{l=1}^2$, and the trials in $Z$ as $\{\mathbf{x}_n\}_{n=3}^N$\;
Query to label the two trials in $S$\;
Construct a linear SVR model $f(\mathbf{x})$ from $S$\;
\For{$k=3,...,K$}{
\For{$n=k,...,N$}{
Compute $d_n^{\mathbf{x}y}$ in (\ref{eq:dnxy})\;}
Identify the $\mathbf{x}'$ that has the largest $d_n^{\mathbf{x}y}$\;
Move $\mathbf{x}'$ from $Z$ to $S$\;
Query to label $\mathbf{x}'$ in $S$\;
Re-index the trials in $S$ as $\{\mathbf{x}_l\}_{l=1}^k$, and the trials in $Z$ as $\{\mathbf{x}_n\}_{n=k+1}^N$\;
Update the linear SVR model $f(\mathbf{x})$ using $S$.}
\uIf{There exist some base estimators which give identical estimates to the true labels in $S$}
{$f(\mathbf{x})$ is the median of these base estimator outputs\;}
\Else{Construct a linear SVR model $f(\mathbf{x})$ from $S$.}
\caption{The AS-iGS active stacking approach.} \label{alg:iGS}
\end{algorithm}

\section{Experiments and Results} \label{sect:experiments}

\subsection{Datasets}

One hundred ECG recordings in the augmented training set of the 2014 PhysioNet/CinC Challenge \cite{Moody2014}, available freely on the PhysioNet platform, were used in this study. They were from patients with a wide range of problems as well as healthy volunteers. Each recording was 10 minutes or shorter, sampled at 360 Hz with 16-bit resolution. Four recordings (2041, 2728, 41024, 41778) shorter than 2 minutes, and one consisting of pure Gaussian noise (42878), were excluded. The remaining 95 ECG recordings had manually annotated QRS complex locations. Heart rates calculated from these locations were used as the references for algorithm evaluations.

\begin{figure}[htpb]\centering
\includegraphics[width=\linewidth,clip]{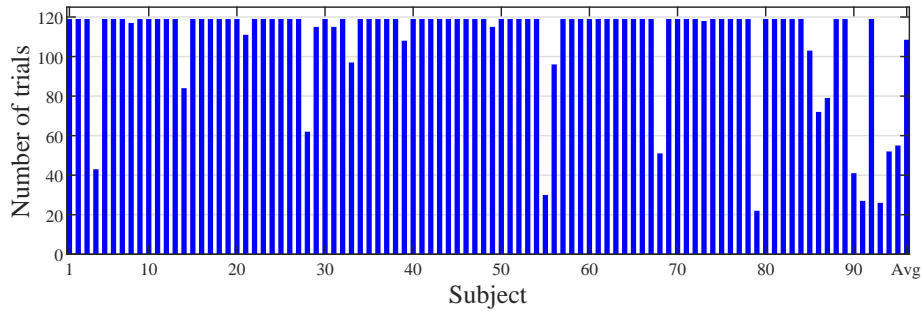}
\caption{Number of trials for the 95 subjects.} \label{fig:numTrials}
\end{figure}

Figure~\ref{fig:numTrials} shows the number of trials from each subject. Most subjects had close to 120 trials, but a few had less than 40. On average each subject had 108.5 trials.

\begin{figure}[htpb]\centering
\includegraphics[width=\linewidth,clip]{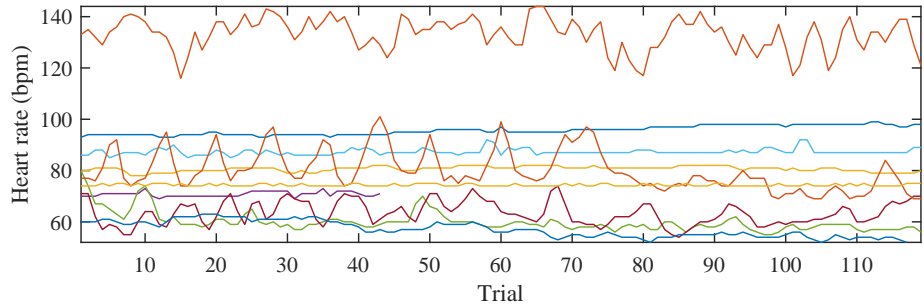}
\caption{Reference heart rates of the first 10 subjects.} \label{fig:heartRate}
\end{figure}

Reference heart rates for the first 10 subjects are shown in Figure~\ref{fig:heartRate}. The heart rates for different subjects differed significantly due to individual differences, and also there may be significant variations within the same subject. These facts make automatic heart rate estimation challenging.

\subsection{Base Estimators}

The following 12 QRS detection algorithms were used as our base estimators \cite{Liu2018}:
\begin{enumerate}
\item Pan-Tompkins \cite{Pan1985}, which has been widely used as a baseline QRS detection algorithm.

\item Hamilton-Tompkins-mean \cite{Hamilton1986}, which is an improvement to the Pan-Tompkins algorithm.

\item Hamilton-Tompkins-median \cite{Hamilton1986}, which is another improvement to the Pan-Tompkins algorithm.

\item RS-slope \cite{Podziemski2013}, which uses the RS slope to detect the QRS complexes.

\item Sixth-power \cite{Dohare2014}, which relies on the sixth power of the ECG signal to identify the QRS complexes.

\item Finite state machine (FSM) \cite{Gutierrez-Rivas2015}, which uses a dynamic finite state machine based threshold to detect the R peaks.

\item Improved FSM (iFSM) \cite{Liu2018}, which improves parameter selection and threshold estimation in FSM.

\item U3 \cite{Paoletti2006}, which uses the U3 transform (a non-linear time-domain transform) for QRS detection.

\item Difference operation algorithm (DOM) \cite{DeCooman2015}, which uses the positive and negative extremes of the low-pass filtered differential ECG signal to detect the R peaks.

\item jqrs \cite{Johnson2015}, which fuses R peaks detected from the ECG using an energy detector with those from the arterial blood pressure waveform using the length transform.

\item Optimized knowledge-based method (OKM) \cite{Elgendi2013}, which detects QRS complexes in ECG signals based on two event-related moving-average filters.

\item UNSW \cite{Khamis2016}, which generates a feature signal containing information of ECG amplitude and derivative, and then performs filtering and adaptive thresholding.
\end{enumerate}

\begin{figure}[htpb]\centering
\includegraphics[width=\linewidth,clip]{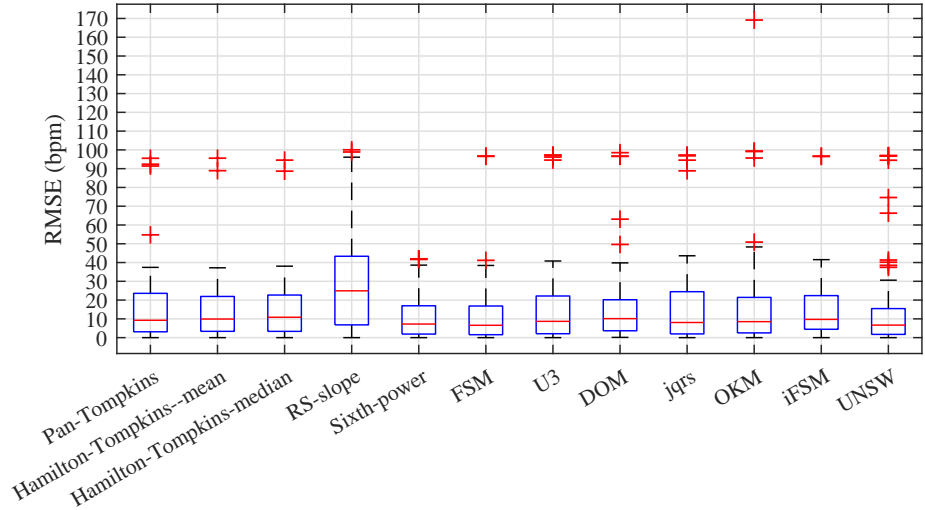}
\caption{Boxplots of the RMSEs of the 12 base estimators on the 95 subjects.} \label{fig:baseLearners}
\end{figure}

\begin{figure}[htpb]\centering
\includegraphics[width=\linewidth,clip]{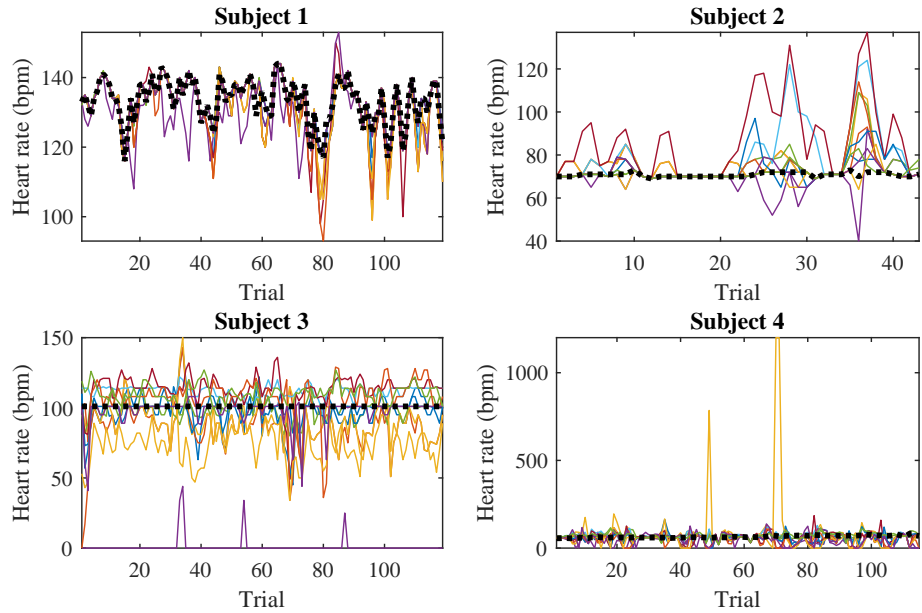}
\caption{Reference heart rates (thick black dotted curve) and the estimates from the 12 base estimators, for four typical subjects.} \label{fig:heartRate12}
\end{figure}

Boxplots of the root mean squared errors (RMSEs) of the 12 base estimators on the 95 subjects are shown in Figure~\ref{fig:baseLearners}. Due to large individual differences, every base estimator broke down on certain subjects, giving heart rate estimates zero or over 1000 bpm, and hence very large RMSEs. Some examples are shown in Figure~\ref{fig:heartRate12}. This is clearly not acceptable in practice. The mean and standard deviation of the RMSEs of the 12 base estimators are shown in the first part of Table~\ref{tab:RMSE}. Among the 12 base estimators, sixth-power achieved the smallest average RMSE (10.55 bpm), and RS-slope the largest (29.57 bpm). Given that the average reference heart rate across the 95 subjects was 87.99 bpm, these RMSEs represented 11.99-33.61\% relative error, suggesting that none of the 12 base estimators can be used alone.

\renewcommand{\baselinestretch}{1}
\begin{table}[htpb] \centering \setlength{\tabcolsep}{4mm} \small
\caption{The mean and standard deviation (std) of the RMSEs of different approaches.}   \label{tab:RMSE}
\begin{tabular}{c|c|c|cc}   \hline
Category   &\multicolumn{2}{c|}{Approach} &RMSE mean (bpm)& RMSE std (bpm)\\ \hline
&\multicolumn{2}{c|}{Pan-Tompkins} & 15.49 & 18.06\\
&\multicolumn{2}{c|}{Hamilton-Tompkins-mean} & 14.69 & 15.78\\
&\multicolumn{2}{c|}{Hamilton-Tompkins-median} & 14.87 & 15.73\\
&\multicolumn{2}{c|}{RS-slope} &29.57 & 26.92\\
&\multicolumn{2}{c|}{Sixth-power} & \textbf{10.55} & \textbf{10.95}\\
Base&\multicolumn{2}{c|}{FSM} & 12.14 & 16.16\\
Estimator&\multicolumn{2}{c|}{iFSM} & 15.26 &16.54\\
&\multicolumn{2}{c|}{U3} & 15.68 &20.47\\
&\multicolumn{2}{c|}{DOM} & 15.67 &19.03\\
&\multicolumn{2}{c|}{jqrs} & 16.33 &20.83\\
&\multicolumn{2}{c|}{OKM} & 17.09 &25.30\\
&\multicolumn{2}{c|}{UNSW} & 14.22 &21.88\\  \hline
Unsupervised&\multicolumn{2}{c|}{LOSO-CV} & \textbf{11.37} &\textbf{11.65}\\
Ensemble&\multicolumn{2}{c|}{Average} & 11.97 &12.14\\
Regression&\multicolumn{2}{c|}{Median} & 12.10 & 16.86\\  \hline
&&RS & 5.55 &4.45\\
&&AS-GSx & \textbf{3.18} &\textbf{3.07}\\
&$K=2$ &AS-RD & 3.76 &4.02\\
&&AS-RD-EMCM & 3.76 &4.02\\
&&AS-iGS & \textbf{3.18} &\textbf{3.07}\\  \cline{2-5}
&&RS & 4.96 & 4.15 \\
&&AS-GSx & \textbf{2.97} & 2.68\\
&$K=3$ &AS-RD & 2.98 & \textbf{2.65} \\
&&AS-RD-EMCM & 3.12 & 2.67 \\
&&AS-iGS & 2.99 & 2.66\\  \cline{2-5}
&&RS & 4.64 &3.97 \\
&&AS-GSx & \textbf{2.81} &\textbf{2.45} \\
&$K=4$ &AS-RD & 2.98 & 2.95 \\
&&AS-RD-EMCM & 3.02 &2.75 \\
Supervised&&AS-iGS & 2.92 &2.78 \\  \cline{2-5}
Stacking&&RS & 4.48&3.89\\
&&AS-GSx & 2.76 &2.70\\
&$K=5$ &AS-RD & \textbf{2.64}&\textbf{2.48} \\
&&AS-RD-EMCM & 2.90 &2.58 \\
&&AS-iGS & 2.99 & 3.05\\  \cline{2-5}
&&RS & 4.40& 3.79\\
&&AS-GSx & \textbf{2.70} &2.69 \\
&$K=6$ &AS-RD & 2.73 & \textbf{2.55} \\
&&AS-RD-EMCM & 2.91 & 2.78 \\
&&AS-iGS & 2.89 & 2.92\\  \cline{2-5}
&&RS & 4.38 & 3.77\\
&&AS-GSx & \textbf{2.66}&2.69 \\
&$K=7$ &AS-RD & 2.82& \textbf{2.66} \\
&&AS-RD-EMCM & 2.91 & 3.00 \\
&&AS-iGS & 2.91 &3.10 \\  \hline
\end{tabular}
\end{table}
\renewcommand{\baselinestretch}{1.5}

Figure~\ref{fig:heartRate12} also shows the reference heart rates for four typical subjects. The base estimators may give many spikes, whereas the reference heart rates were much smoother, suggesting that it is not easy to aggregate the base estimators.

In summary, we have shown that the base estimators were very unstable, and none of them may be used for heart rate estimation alone in practice.

\subsection{Performances of the Unsupervised Ensemble Regression Approaches}

Before testing our proposed active stacking algorithms, we would like to check first if unsupervised ensemble regression can work well. If so, then one should prefer unsupervised ensemble regression, since it does not require manually labeling some ECG trials for each new subject, and hence is very convenient to use.

The following three unsupervised ensemble regression approaches were considered:
\begin{enumerate}
\item \emph{Leave-one-subject-out cross-validation (LOSO-CV)}: From the 95 subjects, each time we selected one as the test subject, and the remaining 94 as training subjects. We combined trials from all 94 training subjects to train a stacking model (we tried both ridge regression with $\lambda=0.01$ and linear SVR; however, the latter was too slow to converge, so we only report the ridge regression results), and computed its RMSE on the test subject. This process was repeated 95 times so that each subject acted as the test subject once. Note that this approach is unsupervised for the new subject, because we do not need any reference heart rates from him/her; however, it assumes that we know the reference heart rates of other subjects, so that the stacking model can be built.

\item \emph{Average}, which has been introduced in Section~\ref{sect:average}.

\item \emph{Median}, which has been introduced in Section~\ref{sect:median}.
\end{enumerate}

\begin{figure}[htpb]\centering
\includegraphics[width=\linewidth,clip]{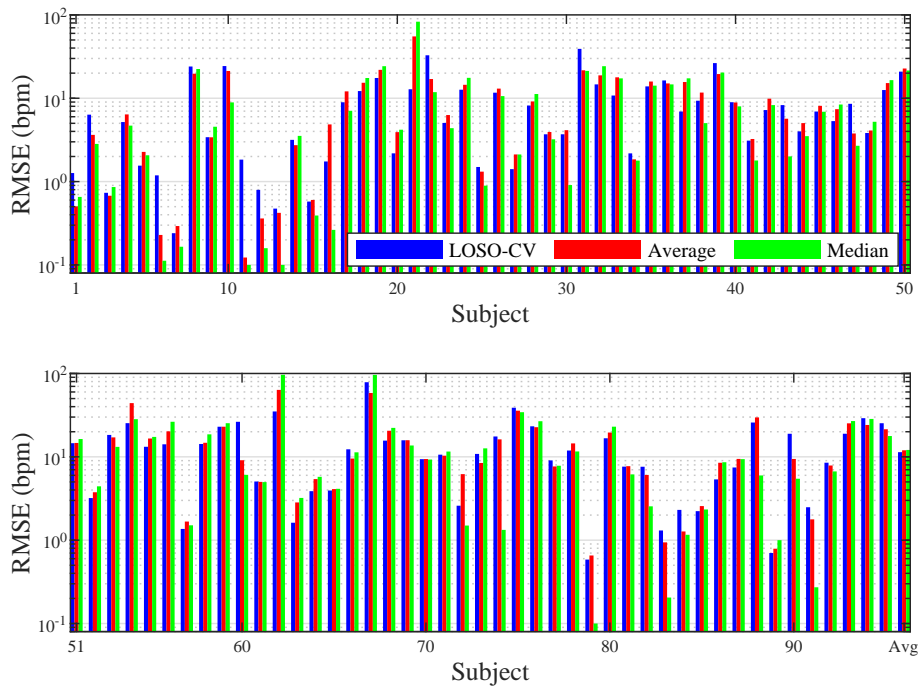}
\caption{RMSEs of the three unsupervised ensemble regression approaches. Logarithmic scale is used for the vertical axis to make the RMSEs more distinguishable.} \label{fig:UER}
\end{figure}

The RMSEs of the three algorithms on the 95 subjects are shown in Figure~\ref{fig:UER}, where the last group in the lower panel shows the mean RMSEs across the 95 subjects. Their values are also shown in the second part of Table~\ref{tab:RMSE}. Given that the mean heart rate across the 95 subjects was 87.99 bpm, these RMSEs represented  $12.92-13.75\%$ relative error, which should not be acceptable in practice.

\begin{figure}[htpb]\centering
\includegraphics[width=.8\linewidth,clip]{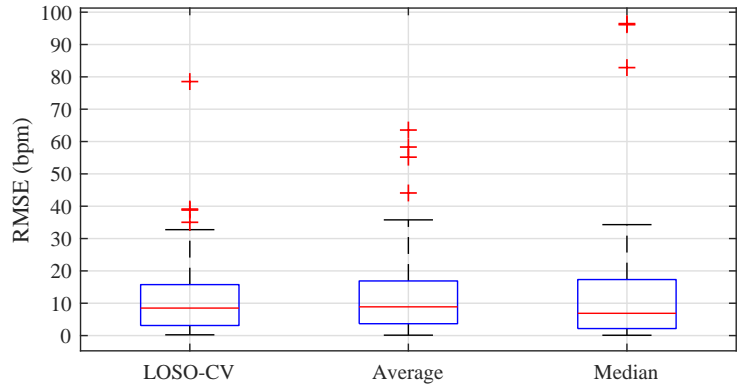}
\caption{Boxplots of the RMSEs of the three unsupervised ensemble regression approaches on the 95 subjects.} \label{fig:boxPlotUER}
\end{figure}

Boxplots of the RMSEs of the three unsupervised ensemble regression approaches on the 95 subjects are shown in Figure~\ref{fig:boxPlotUER}. They were better than most base estimators, but still worse than the best base estimator.

In summary, we have shown that, due to large individual differences, unsupervised ensemble regression approaches may not be accurate enough to be used for practical heart rate estimation.

\subsection{Performances of the Supervised Stacking Approaches}

Next we compare the performances of five supervised stacking algorithms: Random sampling (RS), AS-GSx, AS-RD, AS-RD-EMCM, and AS-iGS. The latter four have been introduced in Algorithms~1-4 in Section~\ref{sect:AS}. RS is similar to AS-GSx, except that GSx is replaced by random sampling.

\begin{figure}[htpb]\centering
\includegraphics[width=\linewidth,clip]{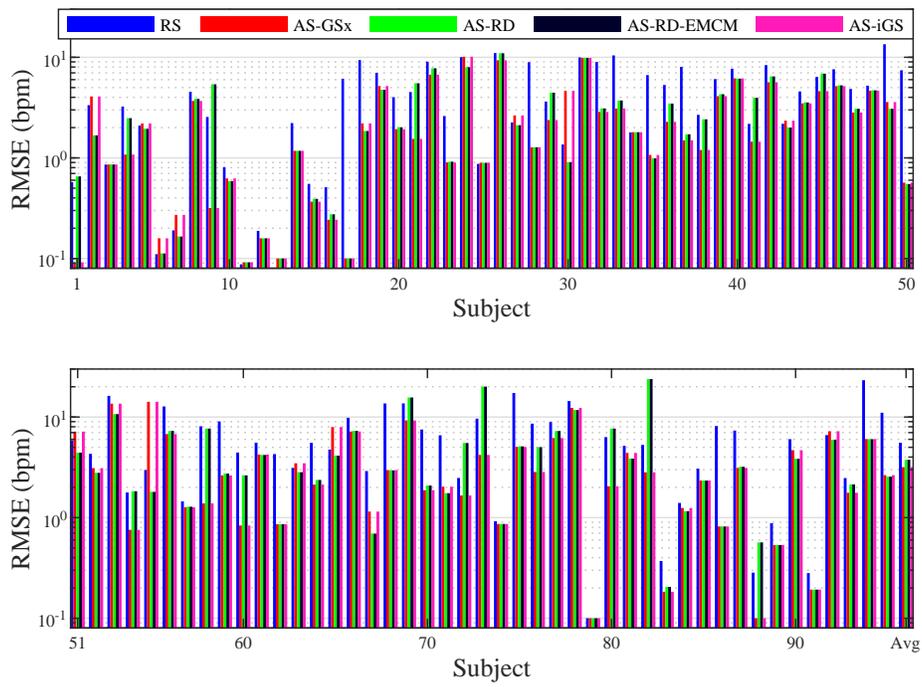}
\caption{RMSEs of the five supervised stacking approaches when $K=2$. Logarithmic scale is used for the vertical axis to make the RMSEs more distinguishable.} \label{fig:AS2}
\end{figure}

\begin{figure}[htpb]\centering
\includegraphics[width=\linewidth,clip]{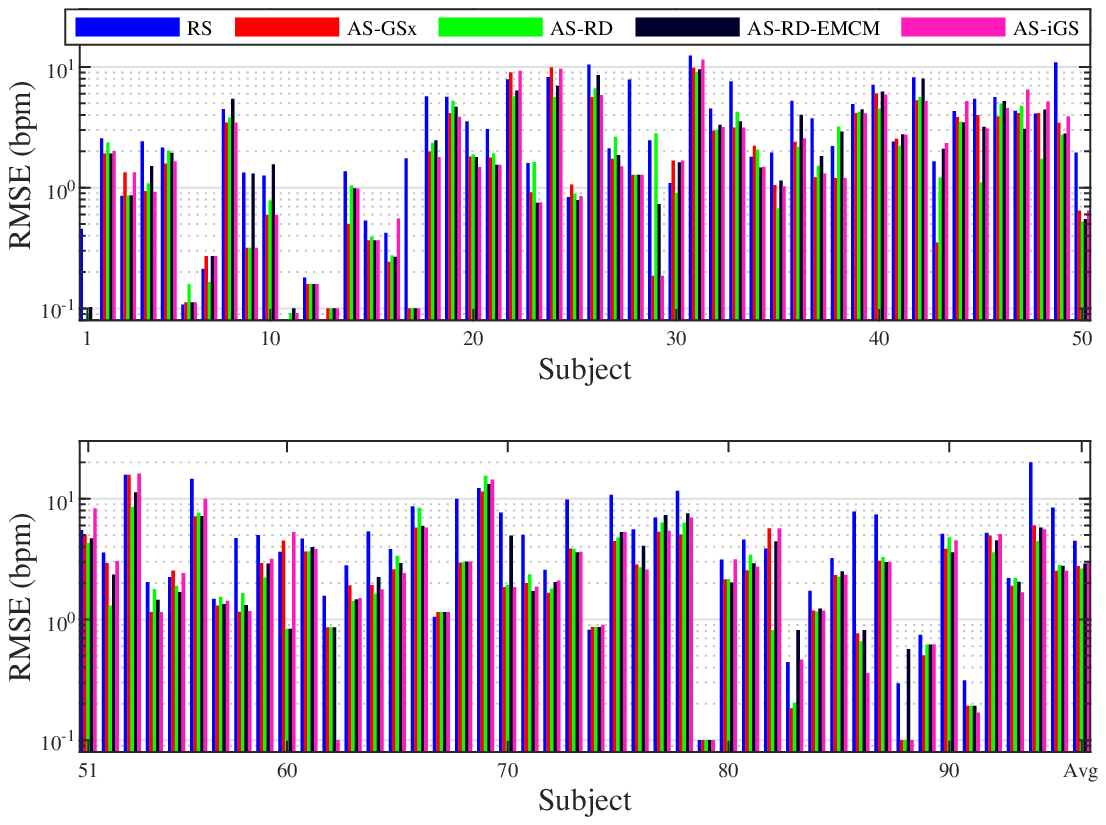}
\caption{RMSEs of the five supervised stacking approaches when $K=5$. Logarithmic scale is used for the vertical axis to make the RMSEs more distinguishable.} \label{fig:AS5}
\end{figure}

The corresponding RMSEs are shown in Figures~\ref{fig:AS2} and \ref{fig:AS5}, respectively, for $K=2$ and $K=5$. Generally, the individual and mean RMSEs were much smaller than those of the three unsupervised ensemble regression approaches (Figure~\ref{fig:UER}).

\begin{figure}[htpb]\centering
\includegraphics[width=\linewidth,clip]{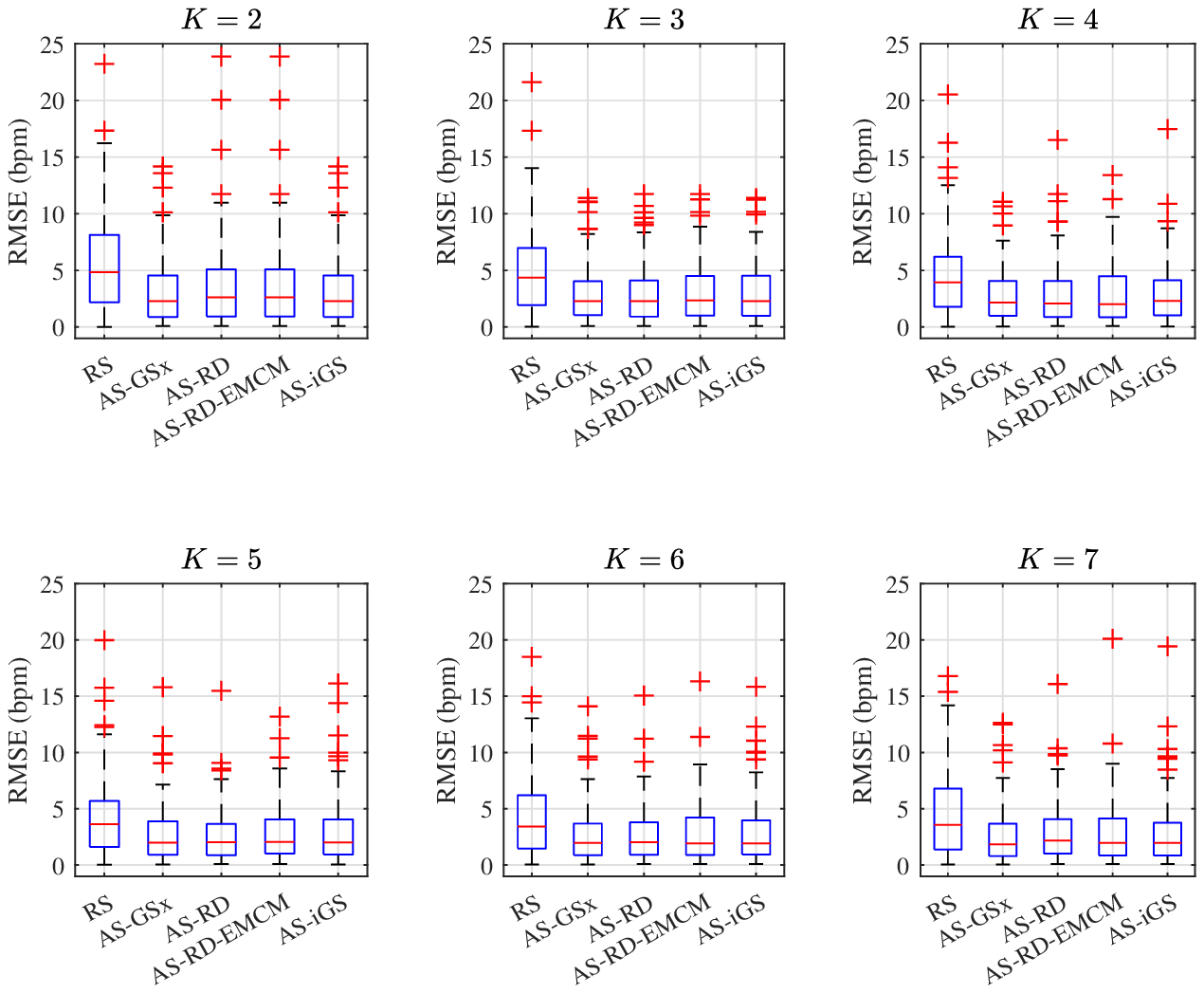}
\caption{Boxplots of the RMSEs of the five supervised stacking approaches, for different $K$.} \label{fig:AS}
\end{figure}

Boxplots of the RMSEs of the five supervised stacking approaches on the 95 subjects are shown in Figure~\ref{fig:AS}, for different $K$. Clearly, these RMSEs were much smaller than those of the 12 base estimators (Figure~\ref{fig:baseLearners}), and also much smaller than those of the three unsupervised ensemble regression approaches (Figure~\ref{fig:boxPlotUER}).

\begin{figure}[htpb]\centering
\includegraphics[width=\linewidth,clip]{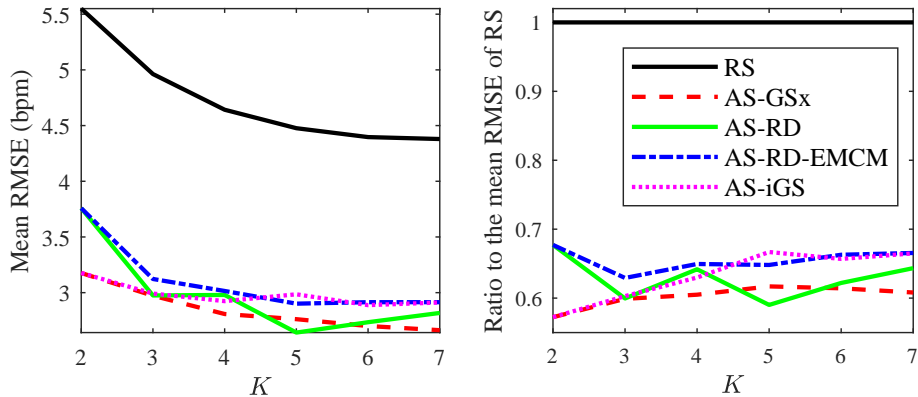}
\caption{Mean RMSEs (left) of the five supervised stacking approaches across the 95 subjects, and the ratio (right) to the mean RMSE of RS.} \label{fig:mRMSE}
\end{figure}

Figure~\ref{fig:AS} also shows that generally the RMSEs of all five supervised stacking approaches decreased with the increase of $K$. To visualize this more clearly, we plot the mean RMSEs of the five supervised stacking approaches across the 95 subjects in the left panel of Figure~\ref{fig:mRMSE}, and also show them in the third part of Table~\ref{tab:RMSE}. Generally there was a decreasing trend for each approach, which is intuitive: the more labeled trials we have, the better a stacking model can be trained. Remarkably, the RMSEs of the four proposed active stacking approaches converged at $K=3$ or $K=4$, i.e., only three or four labeled trials were needed for these active stacking approaches to achieve a low RMSE, which is very favorable in practice.

The left subfigure of Figure~\ref{fig:mRMSE} also shows that the RMSEs of the four active stacking approaches were much smaller than those of RS. The right subfigure of Figure~\ref{fig:mRMSE} shows their ratios to the mean RMSE of RS. Compared with RS, each active stacking approach can reduce the RMSE by $35-40\%$, suggesting the effectiveness of using ALR in heart rate estimation. The four active stacking approaches had similar performances.

To find out if there were statistically significant differences between the five supervised stacking approaches, non-parametric multiple pairwise comparison tests using Dunn's procedure \cite{Dunn1964}, with a $p$-value correction using the False Discovery Rate method \cite{Benjamini1995}, were performed on the $95\times5$ mean RMSEs (for each algorithm on each subject, we computed the mean RMSE for $K\in[2,7]$). The results are shown in Table~\ref{tab:Dunn}, where the statistically significant ones are marked in bold. All four active stacking approaches significantly outperformed RS, but there were no statistically significant differences among the four active stacking approaches.

\begin{table}[htpb] \centering
\caption{$p$-values of non-parametric multiple comparisons on the five supervised stacking approaches $(p=0.05$). } \label{tab:Dunn}
\begin{tabular}{l|cccc}
\hline
     & RS             & AS-GSx           & AS-RD           & AS-RD-EMCM                 \\ \hline
AS-GSx  & \textbf{.0019} &        &                &                    \\
AS-RD & \textbf{.0034}         &.5333 &                & \\
AS-RD-EMCM  & \textbf{.0043} &.5296 &    .4361         &  \\
AS-iGS  & \textbf{.0019} & .4740 & .4858 & .4690  \\ \hline
\end{tabular}
\end{table}

\begin{figure}[htpb]\centering
\includegraphics[width=\linewidth,clip]{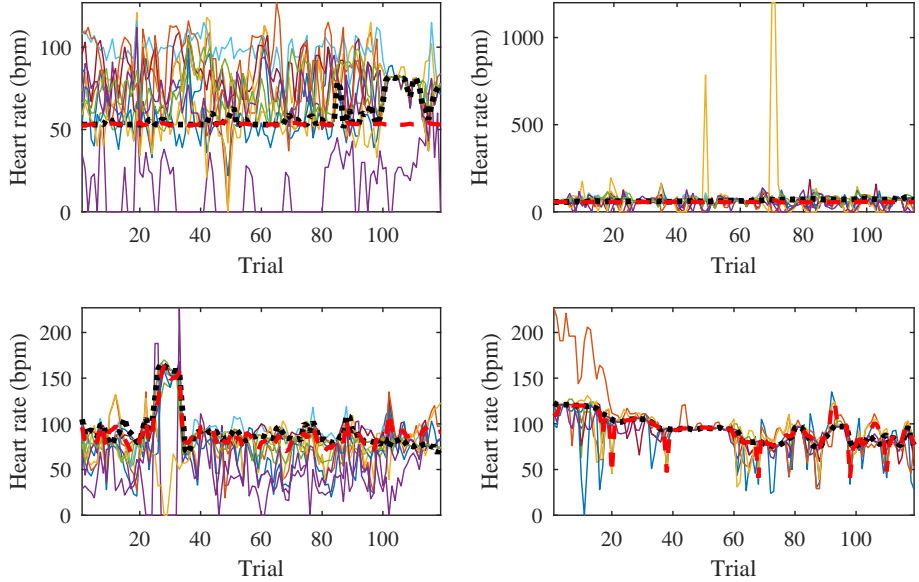}
\caption{Reference heart rates (thick black dotted curves), estimates from AS-GSx (thick red dashed curves), and the estimates from the 12 base estimators, for the four subjects on which AS-GSx ($K=3$) had the largest RMSEs.} \label{fig:AS-GSx}
\end{figure}

Although the four active stacking approaches outperformed the other five approaches, each of them still gave large RMSEs on certain subjects. Figure~\ref{fig:AS-GSx} shows the four subjects on which AS-GSx (with $K=3$) gave the largest RMSEs ($>$10 bpm). For each subject, the 12 base estimators had dramatically different outputs. Clearly, these were very difficult cases.

In summary, we have shown that all five supervised stacking approaches significantly outperformed the 12 base estimators, and the three unsupervised ensemble regression approaches. The four active stacking approaches further significantly outperformed supervised stacking by random sampling. So, active stacking is indeed effective in heart rate estimation.

\subsection{Discussions}

\begin{figure}[htpb]\centering
\includegraphics[width=\linewidth,clip]{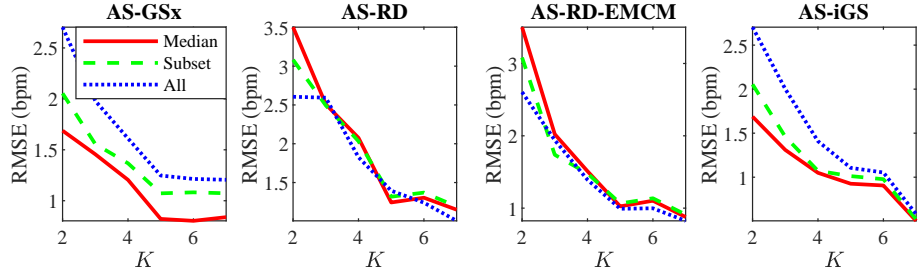}
\caption{Average RMSEs of three variants of the algorithm, when there exist some base estimators whose outputs are identical to the reference heart rates on all selected trials.} \label{fig:median3}
\end{figure}

In all four active stacking approaches (Algorithms~1-4), when there exist some base estimators whose outputs are identical to the reference heart rates on all selected trials, we take the median of these base estimators as the final output, instead of performing a linear SVR. This is because: 1) taking the median is intuitive, as the selected base estimators have identical performance on the reference trials, and hence they cannot be distinguished; 2) taking the median is much simpler than performing a linear SVR; and, 3) empirically taking the median\footnote{We could also take the mean of the selected base estimators; however, it gave a larger RMSE than taking the median, because the mean is more sensitive to outliers than the median.} gave smaller RMSEs. Figure~\ref{fig:median3} shows the average RMSEs of three variants of the algorithm:
\begin{enumerate}
\item \emph{Median}, which takes the median of the selected base estimators.
\item \emph{Subset}, which performs a linear SVR on the selected base estimators.
\item \emph{All}, which performs a linear SVR on all 12 base estimators.
\end{enumerate}
Taking the median had the smallest RMSEs for AS-GSx and AS-iGS, and comparable RMSEs with the two SVR approaches for AS-RD and AS-RD-EMCM (when $K\ge 3$). So, we used the median in our algorithms.

\begin{figure}[htpb]\centering
\includegraphics[width=\linewidth,clip]{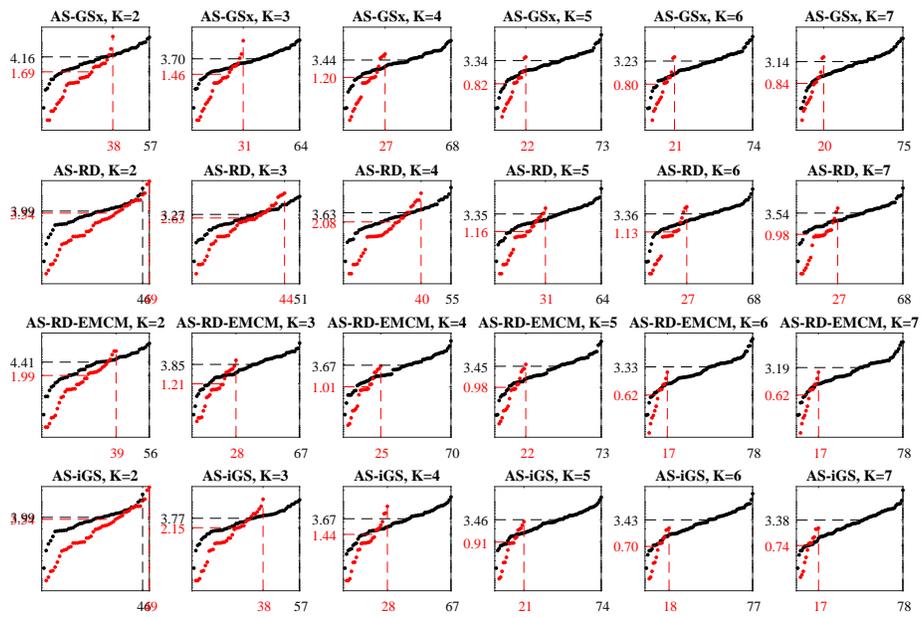}
\caption{Red dots: RMSEs of subjects who have some base estimators whose outputs are identical to the reference heart rates on all $K$ selected trials. Black dots: RMSEs of the remaining subjects. Each dot represents one subject. Dots of the same color are sorted in ascending order for easy visualization. Vertical axis: RMSE in bpm (logarithmic scale is used to better distinguish between the values); horizontal axis: subject. In each subfigure the vertical red (black) dashed line indicates the number of red (black) dots, and the horizontal red (black) dashed line indicates the mean RMSE of the red (black) dots.} \label{fig:median}
\end{figure}

Intuitively, if there exist some base estimators whose outputs are identical to the reference heart rates on all selected trials, then these subjects may be easier to handle than others, i.e., they may have smaller RMSEs. To verify this, we show the RMSEs from these subjects (red dots, sorted in ascending order for easy visualization) versus those from the remaining subjects (black dots, sorted in ascending order for easy visualization) in Figure~\ref{fig:median}. In each subfigure the vertical red (black) dashed line indicates the number of red (black) dots, and the horizontal red (black) dashed line indicates the mean RMSE of the red (black) dots. Each horizontal red line was always lower than the corresponding horizontal black line, confirming our hypothesis. As $K$ increased, the number of red dots decreased (the corresponding vertical red line moved left), which is intuitive, because fewer base estimators were able to completely match the reference heart rates. However, as $K$ increased, the horizontal red line also became lower (the RMSE was smaller), which is reasonable, as the survived subjects were easier to handle.

%
%

\section{Conclusion} \label{sect:conclusion}

Heart rate estimation from ECG signals is very important for the early detection of cardiovascular diseases. However, due to large individual differences and varying ECG signal quality, there does not exist a single reliable estimation algorithm that works well on all subjects. Every algorithm may break down on certain subjects, resulting in a significant estimation error. Ensemble regression, which aggregates the outputs of multiple base estimators for more reliable and stable estimates, is a remedy to this problem. Additionally, active learning can be used to optimally select a few trials from a new subject to label, based on which a stacking ensemble regression model can be trained to properly aggregate the base estimators. This paper has proposed four active stacking approaches, and demonstrated that they all significantly outperformed three common unsupervised ensemble regression approaches, and a supervised stacking approach which randomly selects some trials to label. Remarkably, our active stacking approaches only need three or four labeled trials from each subject to achieve an average root mean squared estimation error below three bpm, making them very convenient for real-world applications. To our knowledge, this is the first research on active stacking, and its application to heart rate estimation.

\section*{Acknowledgement}
This research was supported by the National Natural Science Foundation of China Grants 61873321 and 81871444.


\end{document}